\documentclass[sigconf,screen]{acmart}
\AtBeginDocument{%
  }

\setcopyright{none}
\copyrightyear{2025}
\acmYear{2025}
\acmDOI{XXXXXXX.XXXXXXX}

\makeatletter
\def\@copyrightfooter{%
  \vspace*{\fill}
  \begin{center}
    \footnotesize
    \copyright\ 2024 ACM. Licensed under CC-BY.\\
    ACM-XXXX-YYYY-ZZZZ
  \end{center}
  \vspace{2\baselineskip}%
}
\AtEndDocument{\@copyrightfooter}
\makeatother

\acmConference[MM ’25]{Make sure to enter the correct
  conference title from your rights confirmation email}{October 27--30,
  2025}{Dublin, Ireland}

\usepackage{array}
\usepackage{amsmath}
\usepackage{amsfonts}
\usepackage{colortbl}
\usepackage{arydshln}
\usepackage{multirow}
\usepackage{tabularx} 

\begin{document}

\definecolor{lightyellow}{RGB}{255,255,0}

\title{Generative Sign-description Prompts with Multi-positive Contrastive Learning for Sign Language Recognition}


\author{Siyu Liang}
\email{syliang_233@stu.xidian.edu.cn}
\affiliation{%
  \institution{Xidian University}
  \city{Xi'an}
  \state{Shaanxi}
  \country{China}
}

\author{Yunan Li}
\email{yunanli@xidian.edu.cn}
\affiliation{%
  \institution{Xidian University}
  \city{Xi'an}
  \state{Shaanxi}
  \country{China}
}

\author{Wentian Xin}
\email{wtxin@dlmu.edu.cn}
\affiliation{%
  \institution{Dalian Martime University}
  \city{Dalian}
  \state{Liaoning}
  \country{China}
}

\author{Huizhou Chen}
\email{chenhz@stu.xidian.edu.cn}
\affiliation{%
  \institution{Xidian University}
  \city{Xi'an}
  \state{Shaanxi}
  \country{China}
}

\author{Xujie Liu}
\email{24031110049@stu.xidian.edu.cn}
\affiliation{%
  \institution{Xidian University}
  \city{Xi'an}
  \state{Shaanxi}
  \country{China}
}

\author{Kang Liu}
\email{kangliu@stu.xidian.edu.cn}
\affiliation{%
  \institution{Xidian University}
  \city{Xi'an}
  \state{Shaanxi}
  \country{China}
}

\author{Qiguang Miao}
\authornotemark[1]
\email{qgmiao@xidian.edu.cn}
\affiliation{%
  \institution{Xidian University}
  \city{Xi'an}
  \state{Shaanxi}
  \country{China}
}


\renewcommand{\shortauthors}{Siyu Liang, et al.}

\begin{abstract}

Sign language recognition (SLR) faces fundamental challenges in creating accurate annotations due to the inherent complexity of simultaneous manual and non-manual signals. To the best of our knowledge, this is the first work to integrate generative large language models (LLMs) into SLR tasks. We propose a novel \textbf{G}enerative \textbf{S}ign-description \textbf{P}rompts \textbf{M}ulti-positive \linebreak[2]\textbf{C}ontra\-stive learning (\textbf{GSP-MC}) method that leverages retrieval-augmented generation (RAG) with domain-specific LLMs, incorporating multi-step prompt engineering and expert-validated sign language corpora to produce precise multipart descriptions. The GSP-MC method also employs a dual-encoder architecture to bidirectionally align hierarchical skeleton features with multiple text descriptions (global, synonym, and part level) through probabilistic matching. Our approach combines global and part-level losses, optimizing KL divergence to ensure robust alignment across all relevant text-skeleton pairs while capturing both sign-level semantics and detailed part dynamics. Experiments demonstrate state-of-the-art performance against existing methods on the Chinese SLR500 (reaching 97.1\%) and Turkish AUTSL datasets (97.07\% accuracy). The method's cross-lingual effectiveness highlight its potential for developing inclusive communication technologies.

\textcolor{red}{This work has been submitted to the IEEE for possible publication. Copyright may be transferred without notice, after which this version may no longer be accessible.}
  
\end{abstract}


\begin{teaserfigure}
  \centering
  \includegraphics[width=0.9\textwidth]{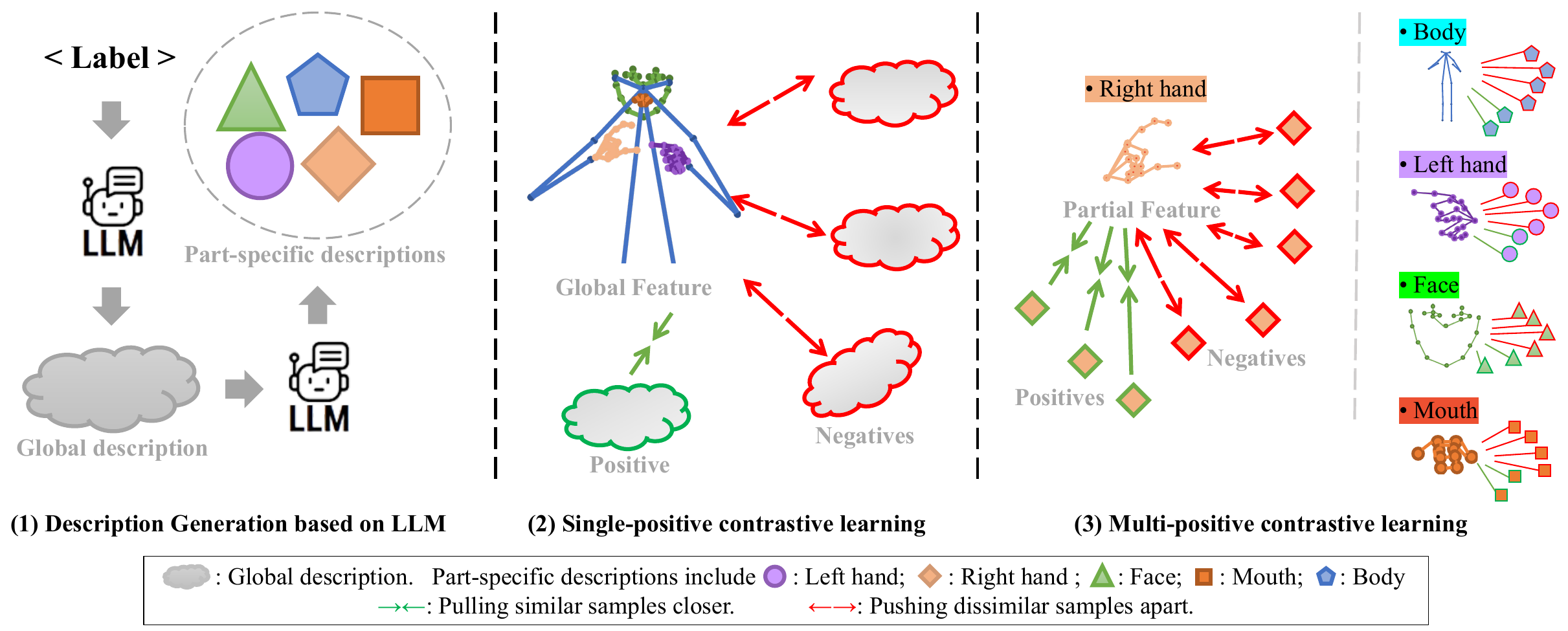}
  \caption{Multi-positive contrastive learning for LLMs generated multipart sign-description. }
  \Description{Multi-positive contrastive learning for LLM-generated multipart sign-description.}
  \label{fig:multi-positive}
\end{teaserfigure}


\maketitle

\section{Introduction}
Due to its importance in bridging communication gaps across diverse human communities, sign language recognition (SLR) has attracted substantial research interest. The advent of high-precision depth sensors, including Kinect \cite{MicrosoftKinect2017} and RealSense \cite{IntelRealSense2019}, coupled with advances in pose estimation algorithms \cite{fang2022alphapose, sun2019deep}, , has significantly simplified the acquisition of human joint positions in recent years. Skeleton-based SLR methods, which rely on body joint movements, have gained considerable attention due to their computational efficiency and robustness to variations in lighting conditions, viewpoints, and background noise. An increasing number of SLR works are based on the skeleton modality \cite{lin2024skim} or use skeletons as one of the multimodal inputs \cite{jiao2023cosign, zhao2023best, zhao2024masa, liang2025integrated}.

The remarkably rapid development of pre-trained generative LLMs has led to their expanding applications across various recognition domains \cite{yuan2022skeletonclip, bao2020unilmv2, liu2025sg}, particularly in action recognition. Approaches such as GAP \cite{xiang2023gap} and SpaMo \cite{hwang2024efficient} have demonstrated how LLMs can enhance recognition by generating fine-grained textual descriptions or interpreting spatial-motion features. Although these methods primarily employ LLMs as sophisticated text processors, their generative capabilities remain underexplored for domain-specific applications like sign language recognition. This presents unique challenges, as sign language processing requires both linguistic precision and specialized domain knowledge that general-purpose LLMs typically lack. Potential solutions must address the fundamental tension between domain-specific motion accuracy and linguistic expressiveness in generated descriptions.

Although prompt engineering has enabled LLMs to assist in action recognition by generating auxiliary text descriptions, such approaches face unique hurdles in SLR. Sign language requires expert knowledge, as subtle variations in hand shape, motion, or expression convey different meanings. General LLMs often produce descriptions that are either overly generic or semantically inconsistent. Existing methods \cite{xiang2023gap, yan2024crossglg}, designed for action recognition,  struggle with inaccuracies and hallucinations in sign descriptions. This limitation calls for domain-aware prompting techniques that harmonize LLMs' generative flexibility with the structural precision demanded by sign language. This suggests the need for new prompting paradigms that can bridge the generative capacity of LLMs with the strict descriptions of expert-defined signs.

Contrastive learning has revolutionized unsupervised representation learning in multiple domains. MoC \cite{he2020momentum} implements a momentum-based dynamic dictionary with a queue and moving-average encoder for scalable contrastive learning. SimCL \cite{chen2020simple} proposes a me\-thod for text-to-image models, treating multiple images generated from the same text prompt as positive samples. MC \cite{liu2025ecl} proposes a multi-view enhanced contrastive learning method for visual representation by maximizing agreements between multi-view radiographs and their corresponding reports, solving medical imaging issues.
Sign language recognition challenges contrastive learning with many-to-many relationships: One sign has multiple valid descriptions (all of which should be treated as positives), and descriptions often focus on partial actions. The single positive contrastive learning method struggle with these variable, incomplete positives, requiring new approaches that handle probabilistic alignments while preserving discrimination.

The contributions of our proposed method are as follows.

1. To the best of our knowledge, we are the first to integrate generative LLMs into SLR through our \textbf{G}enerative \textbf{S}ign-description \textbf{P}rompts (GSP). GSP employs retrieval-augmented generation with domain-specific LLMs to produce accurate multipart sign descriptions, providing reliable textual grounding for contrastive learning.

2. We design the \textbf{M}ulti-positive \textbf{C}ontrastive learning (MC) approach, which combines retrieval-augmented generative descriptions from expert-validated knowledge bases and a novel multi-positive contrastive learning paradigm.

3. Comprehensive experiments on the Chinese SLR500 and Turkish AUTSL datasets further  validate the effectiveness of our method, achieving state-of-the-art accuracy (97.1\% and 97.07\% respectively). The consistent and robust performance across languages demonstrates generalization capabilities.

\begin{figure*}[ht]
  \centering
  \includegraphics[width=0.8\linewidth]{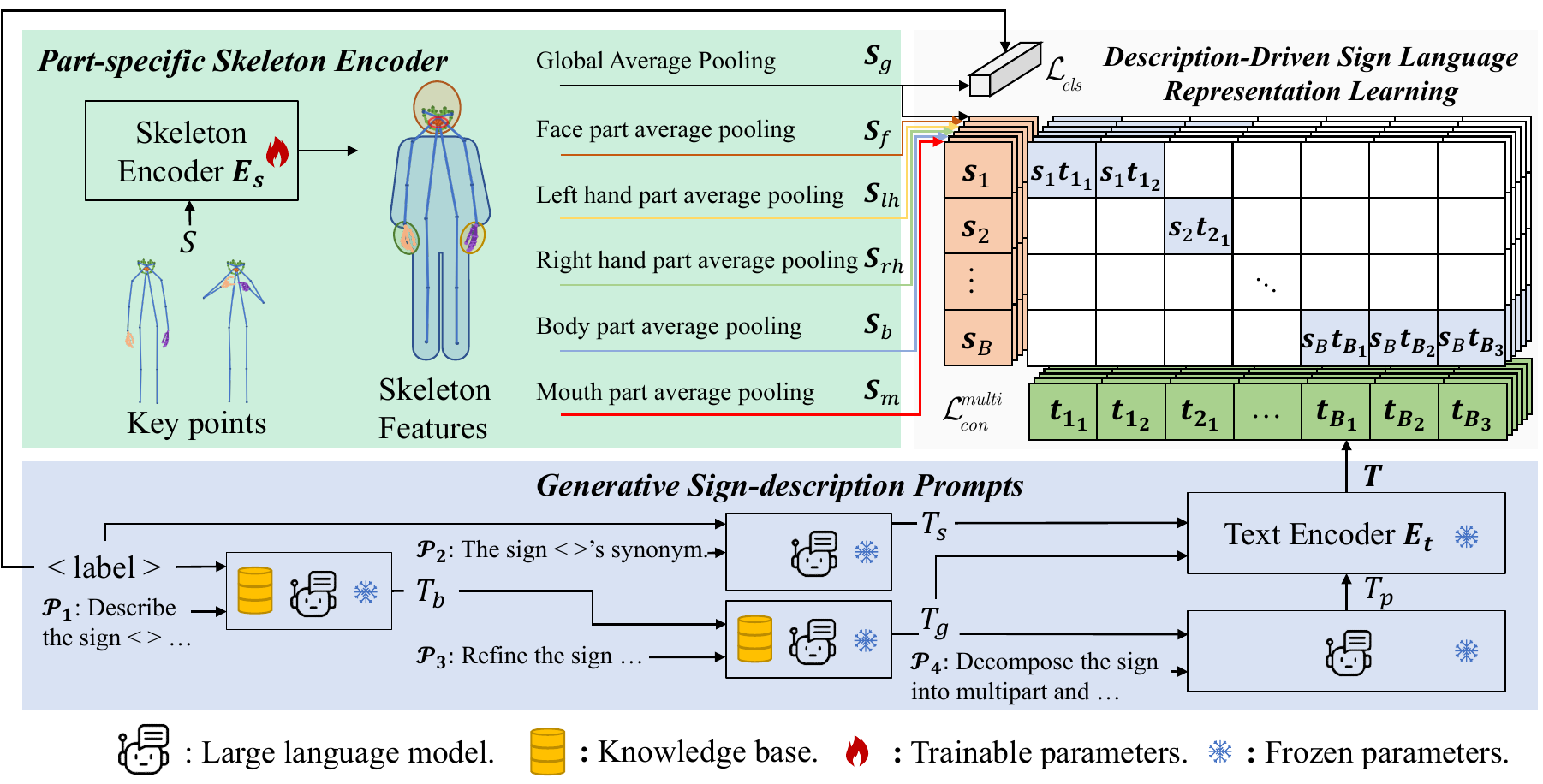}
  \caption{The overall architecture of GSP-MC method. $\mathbf{S}_g$ denotes global skeleton features for classification, $\mathbf{S}_f$, etc. represents part-specific features for contrastive learning, and $\mathbf{T}$ is encoded text features. Training is guided by $\mathcal{L}_{cls}$ and $\mathcal{L}_{con}$ losses.}
  \label{fig:framework}
  \Description{The framework of proposed method. It shows the architecture with components for skeleton feature extraction, text description generation, classification learning, and multimodal representation alignment.}
\end{figure*}
\section{Related Works}

\subsection{Skeleton-Based Sign Language Recognition}

Current research in skeleton-based SLR has explored several promising approaches. Self-supervised learning approaches, such as Sign-BER \cite{hu2021signbert} and BES \cite{zhao2023best}, employ masked sequence modeling and quantized pseudo-labels to learn sign representations from hand pose sequences. These methods demonstrate the potential for learning sign representations without extensive annotations, although they primarily focus on low-level motion patterns rather than linguistic semantics. Graph-based architectures like CoSig \cite{jiao2023cosign} utilize specialized graph convolutional networks to capture the skeletal dynamics, but often neglect the global semantic relationships inherent in the sign language. Meanwhile, multimodal pretraining methods such as MAS \cite{zhao2024masa} integrate motion-aware autoencoders with semantic alignment to learn joint visual-linguistic representations. Despite these advances, domain methods exhibit persistent limitations by treating sign language primarily as motion sequences while neglecting its structured linguistic nature. This leads to excessive dependence on data-driven motion patterns, weak integration with linguistic knowledge, and inadequate visual-semantic alignment, which are fundamental shortcomings that hinder linguistically grounded SLR systems. Thus, our work addresses these gaps by multi-positive contrastive learning, aligning each sign with multiple expert-verified descriptions while maintaining cross-modal discrimination via fixed-text-encoder training. This approach learns robust visual features that absorb natural sign variations without computational overhead during inference, achieving more linguistically-grounded recognition than previous methods.

\subsection{Text-Augmented Sign Language Representation}

Recent progress in combining textual and visual features for SLR predominantly adopt two distinct strategies. On the one hand, manual text encoding methods such as NLA-SLR~\cite{zhuo2023nla} and $C^2ST$~\cite{zhang2023c2st} leverage predefined gloss vocabularies, but are constrained by static description sets that fail to capture execution variations and face scalability issues due to high manual annotation costs. On the other hand, generative text integration approaches, including Action-GPT~\cite{kalakonda2023action} and GAP~\cite{xiang2023gap}, employ LLM-based action description generation. However, these primarily target generic action recognition or vision-language tasks, leaving key SLR challenges unaddressed: hallucinated sign descriptions, lack of sign-specific linguistic constraints, and difficulties in modeling simultaneous sign components (e.g., hand shape, motion, and orientation). Notably, while Sign2GPT~\cite{wong2024sign2gpt} attempts integrate LLM, current methods still lack robust generative LLM solutions for these domain-specific requirements. Our proposed approach tackles these issues through RAG to anchor LLM outputs in knowledge bases and multi-level visual-text alignment constrained by various generated descriptions.

\section{Methods}

The Generative Sign-description Prompts with Multi-positive Contrastive Learning (GSP-MC) method augments skeleton-based sign language recognition through multimodal representation learning. As illustrated in Figure~\ref{fig:framework}, the architecture comprises three elements: (1) a group-specific skeleton encoder extracting hierarchical motion features, (2) a generative sign-description prompts (GSP) module producing expert knowledge grounded text descriptions, and (3) a multi-positive contrastive learning mechanism( MC) aligning visual and textual representations. The method maintains computational efficiency during inference using only the skeleton encoder.

The skeleton encoder $E_s$ extracts both the global skeleton features $\mathbf{S}_g$ and the part-specific features $\mathbf{S}_p$ from the input pose sequences. The GSP employs LLMs to produce descriptive texts $t$, which are subsequently encoded into text features $\mathbf{T}$ by the text encoder $E_t$. These multimodal features are then optimized through our proposed multi-positive contrastive learning approach, which enhances the model's capacity to capture fine-grained action semantics from textual descriptions.

\begin{table*}
\centering
\renewcommand{\arraystretch}{1.8}
\caption{Alignment results comparing primary $T_b$ and refined $T_g$ descriptions, highlighting: \textcolor{blue}{sign label}, \textcolor{red}{substitute descriptions}, \textcolor[rgb]{0.8,0.8,0}{expert-validated knowledge}, primary descriptions \colorbox[rgb]{0.8549, 0.8902, 0.9529}{$T_b$}, and refined descriptions \textcolor[rgb]{0.3294,0.8698,0.2078}{$T_g$}.}
\label{tab:alignment-results}
\begin{tabularx}{\textwidth}{
  >{\hsize=0.9\hsize}X  
  >{\hsize=0.9\hsize}X  
  >{\hsize=1.2\hsize}X  
}
\hline
\multicolumn{1}{c}{\textbf{ Primary descriptions $T_b$}} & 
\multicolumn{1}{c}{\textbf{Expert-validated knowledge}} & 
\multicolumn{1}{c}{\textbf{ Refined descriptions $T_g$}} \\
\noalign{\vskip-1pt}\hline\noalign{\vskip1pt}
\cellcolor[rgb]{0.8549, 0.8902, 0.9529} \textcolor{blue}{Devoted: }(I) \textcolor{red}{Make the sign for ``love''.} (II) Extend the thumb with one hand and place it on the palm of the other hand, then raise it upwards. & 
The sign for ``love'': \textcolor[rgb]{0.8, 0.8, 0}{Gently caress the back of the thumb with one hand, expressing a feeling of ``tenderness''.} & 
\textcolor[rgb]{0.3294, 0.8698, 0.2078}{\textcolor{blue}{Devoted: }(1) Gently caress the back of the thumb with one hand, expressing a feeling of ``tenderness''. (2) Extend your thumb with one hand, sit on the other palm, and lift it up.} \\
\noalign{\vskip-1pt}\hline\noalign{\vskip1pt}
\cellcolor[rgb]{0.8549, 0.8902, 0.9529} \textcolor{blue}{Ambience:} (1) One hand \textcolor{red}{forms the manual sign ``Q''}, with the fingertips pointing inward, placed at the nostrils. (2) Extend your index finger with one hand and make a big circle with your fingertips facing down. & 
The manual sign ``Q'': \textcolor[rgb]{0.8, 0.8, 0}{One hand with the right thumb down, the index and middle fingers together on top, the thumb, index, and middle fingers pinched together, the fingertips pointing forward and slightly to the left, the ring and little fingers bent, the fingertips touching the palm.} & 
\textcolor[rgb]{0.3294, 0.8698, 0.2078}{\textcolor{blue}{Ambience:} (1) One hand with the right thumb down, the index and middle fingers together on top, the thumb, index, and middle fingers pinched together, the fingertips pointing forward and slightly to the left, the ring and little fingers bent, the fingertips touching the palm. The fingertips are pointing inward, placed at the nostrils. (2) Extend index finger with one hand and make a circle with your fingertips facing down.} \\
\hline
\end{tabularx}
\end{table*}

\subsection{Part-specific Skeleton Encoder}

Our model processes input skeleton sequences $S \in \mathbb{R}^{B \times 3 \times N \times T}$ as input, where $B$ is the batch size, $3$ represents (x,y,confidence), $N$ denotes the number of joints, and $T$ indicates the temporal sequence length. The model predicts the labels $l \in \mathbb{R}^{B1}$ as output.

\textbf{Keypoint Selection and Grouping.} Using HR-Net~\cite{sun2019deep} as our skeleton estimator, we extract 87 keypoints per frame, strategically divided into five anatomical groups: 15 for the body $S_b$, 21 for each hand $S_{lh}$ and $S_{rh}$), 10 for the mouth $S_m$ and 20 for the face $S_f$, respectively. We provide a detailed analysis of this keypoint selection strategy in Section~\ref{sec:keypoint-selection}.

\textbf{Skeleton Encoder.} Representing the skeleton as a graph $S = \{V, E\}$ with joints $V$ and edges $E$, we process each part $P$ through layered graph convolutions.

\begin{equation}
\mathbf{S}_{P,l+1}=\sigma\left(\mathbf{D}^{-\frac{1}{2}}\mathbf{A}\mathbf{D}^{-\frac{1}{2}}\mathbf{S}_{P,l}\Theta_{l}\right)
\end{equation}
where, $\mathbf{D} \in \mathbb{R}^{N \times N}$ is the degree matrix, $\mathbf{A}$ the adjacency matrix , $\Theta_l \in \mathbb{R}^{C_l \times C_{l+1}}$ the learnable parameter at layer $l$, and $\sigma$ the activation function.

Our basic block employs multiple CTR-GC block \cite{chen2021channel}, with each block integrating a Channel-wise Topology Refinement Graph Convolution layer. We fuse channel-wise topology refinement with graph convolution to yield the final part-specific representation.

\textbf{Skeleton Classification.} 
The model optimizes a standard cross-entropy loss:

\begin{equation}
\mathcal{L}_{cls} = -\mathbf{Y}\log p_{\theta}(\mathbf{S}_g)
\end{equation}
where $\mathbf{Y}$ denotes ground truth labels, $\mathbf{S}_g$ the global skeleton features, and $p_{\theta}(x)$ the predicted probability distribution.

\subsection{Generative Sign-description Prompts (GSP)}
\label{sec:generative-sign-description-prompts}

\begin{figure}[ht]
  \centering
  \includegraphics[width=\linewidth]{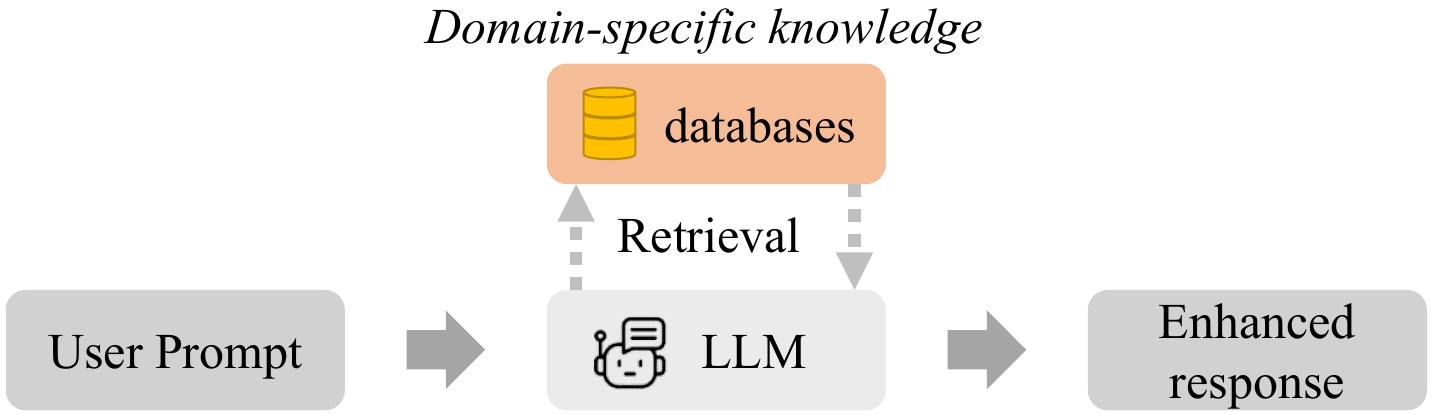}
  \caption{Architecture of the customized LLM for sign description generation.}
  \label{fig:custom-llms}
  \Description{The workflow of custom LLMs.}
\end{figure}

The Generative Sign-description Prompts (GSP) establishes a systematic approach for generating accurate and comprehensive sign language descriptions through advanced language model techniques. As illustrated in Figure~\ref{fig:custom-llms}, the system integrates domain-specific knowledge from authoritative sources including official sign definitions and expert textbooks, with a capacity supporting up to 1GB of domain-specific data and individual documents up to 50MB (~15 million Chinese characters). This extensive knowledge base serves as the foundation for addressing key challenges in sign language description generation.

\subsubsection{Expert Dataset Construction and Standardization}

The method addresses the inherent challenges of professional sign language descriptions through a dual-path generation mechanism. Professional sign language definitions frequently employ substituted descriptions that create barriers for machine interpretation, manifesting as standardized manual alphabet references (``uses the `5' handshape'') or cross-sign equivalencies (``identical to the sign `good'{''}). To overcome this, the system implements a rigorous standardization process where primary descriptions $T_b$ are generated through RAG-augmented generation anchored in authoritative sources:

\begin{equation}
T_b = \text{LLM}_{\text{RAG}}(\text{label}, \mathcal{P}_1)
\end{equation}

Simultaneously, the method incorporates controlled diversity through synonym variations $T_s$ generated by leveraging the LLM's creative potential within carefully designed constraints:

\begin{equation}
T_s = \text{LLM}(\text{label}, \mathcal{P}_2)
\end{equation}

The prompt templates $\mathcal{P}_{1,2}$ incorporate domain-specific instruction tuning to ensure outputs maintain both factual accuracy through expert grounding and linguistic diversity. For instance, a base description ``palm pushes forward'' might be expanded to ``arm extends with palm facing outward'' while preserving core features. This dual method effectively balances the need for standardization against the requirement for varied expressions to enhance model robustness.

\subsubsection{Redundancy Elimination and Multi-part Decomposition}

The method implements advanced processing to address extraneous information common in professional sign language materials. Through targeted prompt design $\mathcal{P}_4$, the system automatically filters irrelevant content such as homophonic explanations ({``}`One' is homophonous with `idea'{''}), focusing exclusively on action-relevant descriptions. As show in Table \ref{tab:alignment-results}, the refinement process transforms initial outputs $T_b$ into complete  formulations $T_g$ while preserving expert-validated knowledge:

\begin{equation}
T_g = \text{LLM}_{\text{RAG}}(T_b, \mathcal{P}_3)
\end{equation}

Furthermore, the system introduces innovative multi-part decomposition, automatically generating part-specific texts $T_p$ with corresponding anatomical annotations:

\begin{figure}[ht]
  \centering
  \includegraphics[width=\linewidth]{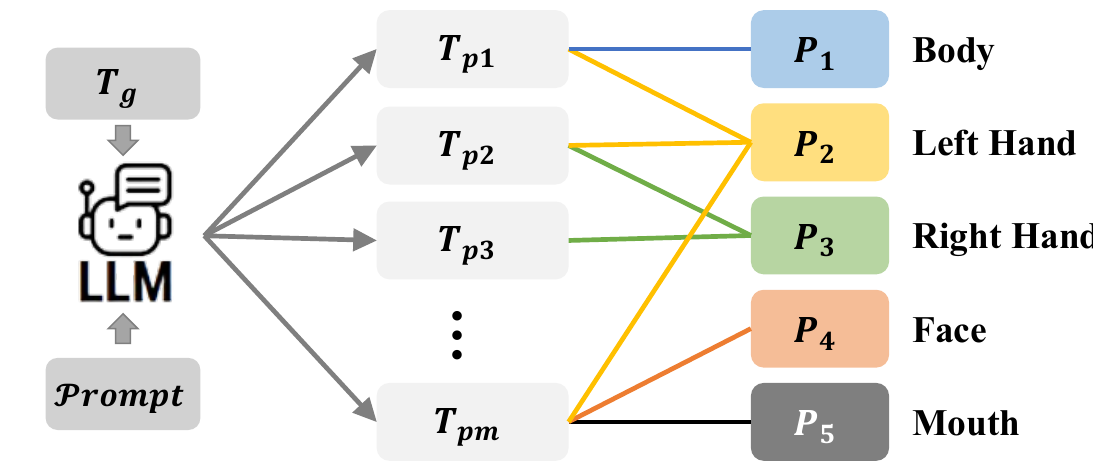}
  \caption{Generation of multipart texts.}
  \label{fig:multipart-text-generate}
  \Description{Generation of multipart texts.}
\end{figure}

\begin{equation}
T_p = \text{LLM}(T_g, \mathcal{P}_4)
\end{equation}

The structured output $T_p$ pairs each text segment ($T_{p1}$, $T_{p2}$, etc.) with its corresponding sign language components. As demonstrated in Figure~\ref{fig:multipart-text-generate}, $T_{p1}$ describes the movement involving both $part_1$ and $part_2$, while $T_{p2}$ relates to $part_2$ and $part_3$, which establish a many-to-many mapping between text and human body parts.

\subsubsection{Text Encoder}

The text encoding component employs the CLIP text encoder as its backbone $E_t$, processing three distinct text modalities: global descriptions $T_g$, semantically-rich synonym variants $T_s$, and fine-grained part-specific texts $T_p$. The encoding process begins with standard tokenization, followed by processing through 12 transformer blocks with multi-head self-attention mechanisms. This architecture generates context-aware hierarchical linguistic representations that are efficiently aggregated into fixed-dimensional feature vectors $\mathbf{T_g}$, $\mathbf{T_s}$, and $\mathbf{T_p}$.

For part-specific encoding, individual body part descriptions undergo independent processing through identical architectural components, ensuring feature space consistency across all modalities. The use of frozen pre-trained weights maintains established linguistic knowledge while providing computational efficiency during training. This design choice allows the method to focus innovation on the description generation aspects while leveraging proven text encoding capabilities.

\subsection{Description-Driven Sign Language Representation Learning}

\subsubsection{Text-Conditioned Multi-Positive Alignment}

Our method uses a dual-encoder architecture. It includes a skeleton encoder $E_s$ and a text encoder $E_t$. $E_s$ processes skeletal motion data $\mathbf{S} \in \mathbb{R}^{(B \times 3 \times N \times T)}$, while $E_t$ handles action descriptions. Unlike traditional one-hot label supervision, the proposed approach uses natural language descriptions, which provide richer supervision for skeleton classification. As illustrated in Figure~\ref{fig:multi-positive}, our method simultaneously aligns each skeleton sample with multiple relevant text descriptions while maintaining separation from negative samples.

In the global description scenario, where each skeleton sequence corresponds to one global description $T_g$ or one synonym variant $T_s$, the number of text features $M$ exactly matches the batch size $B$ (i.e., $M=B$). This configuration reduces to conventional single-positive contrastive learning, with each skeleton feature contrasted against its single paired text feature.

In the part-specific description scenario, the method demonstrates its full capability when processing part-specific descriptions $T_p$. Here, multiple textual descriptions (ranging from 0 to $m$ per body part) are generated for each skeleton sequence, resulting in the total text features of $M=\sum_{i=1}^B m_i$, where $M>B$. This expanded correspondence enables genuine multi-positive contrastive learning, as each skeleton feature can simultaneously align with multiple relevant text descriptions capturing different aspects of the same sign action. While $T_g$ and $T_s$ maintain the $M=B$ correspondence, the part-specific $T_p$ descriptions create the $M>B$ scenario that drives our multi-positive learning advantage. This hierarchical text representation allows the model to simultaneously learn both holistic sign semantics and fine-grained part dynamics.

The dual-encoders are jointly optimized by contrasting skeleton-text pairs in two directions within the batch:

\begin{equation}
\mathbf{q}_{ia}^{s \to t} (\mathbf{s}) = \frac{\exp(\operatorname{sim}(\mathbf{s}_i, \mathbf{t}_{ij}) / \tau)}{\sum_{i=1}^{B}\sum_{k=1}^{m} \exp(\operatorname{sim}(\mathbf{s}_i, \mathbf{t}_{ik}) / \tau)}
\end{equation}
\begin{equation}
\mathbf{q}^{t \to s} (\mathbf{t}) = {\mathbf{q}^{s \to t} (\mathbf{t})}^T
\end{equation}
where $s$, $t$ are encoded features of skeleton and text. $a=\sum_{b=1}^{i-1}m_b+j$. $j$ represents the index of the $j$-th text feature within the set of $m$ text features that correspond to the $s_i$ keypoint action feature.
$sim(s, t)$ is the cosine similarity, $\tau$ is the temperature parameter and $B$ is the batch size. Unlike image-text pairs in CLI \cite{radford2021learning}, which are one-to-one mappings, in our setting, there could be more than one positive matching and actions of different categories forming negative pairs. 

Given multiple candidate text features for each sign language sample, we establish a probabilistic match where at least one text feature $t$ corresponds to each skeleton feature $s$. The true correspondence distribution $\mathbf{p}\in\mathbb{R}^{(B,M)}$ is defined as:

\begin{equation}
\mathbf{p}_{i}=\frac{\mathbb{I}_{\mathrm{m a t c h} ( s, t_{i} )}} {\sum_{c=1}^{M} \mathbb{I}_{\mathrm{match} ( s, t_{c} )}}
\end{equation}
where the indicator function $\mathbb{I}_{\mathrm{match}(s,t_{i})}$ equals 1 when $s$ and $t_i$ share the same sign label, and 0 otherwise. This formulation explicitly captures the multi-positive relationships between skeleton features and their corresponding text descriptions.

The contrastive learning objective employs KL divergence to align the predicted and true distributions:
\begin{equation}
\mathcal{L}_{\text{con}}(s,t) = \frac{1}{2} \mathbb{E}_{\mathbf{s, t} \sim \mathcal{D}} \left[ \text{KL}(q^{s \to t}(\mathbf{s}), \mathbf{p}) + \text{KL}(q^{t \to s}(\mathbf{t}),\mathbf{p}^{T}) \right]
\end{equation}
where $D$ is the entire dataset. 

This symmetric loss function generalizes standard single-positive contrastive learning \cite{oord2018representation}, where $\mathbf{p}$ reduces to a one-hot vector. Although related to \cite{tian2023stablerep}, our key distinction is the explicit treatment of all texts of the same label as matched positives. Optimization brings together partial descriptions of identical signs while separating them from different signs, enhancing feature discriminability.

\subsubsection{Hierarchical Part Contrastive Learning}

Considering the prior of human body parts, the skeleton can be divided into multiple groups. We illustrate the overall architecture in Figure \ref{fig:framework}. We apply contrastive loss on different part features as well as global feature, and propose a multipart contrastive loss. The part feature could be obtained with part pooling, where joint features within the same group are aggregated to generate a part representation. More specifically, we choose the features before the final classification layer for part feature pooling. Our part partition strategy is five-part partition. The body is divided into five groups: left hand, right hand, face, mouth, and body. We then engage in contrastive learning by comparing the sign-relevant partial text features, obtained from section \ref{sec:generative-sign-description-prompts}, with their corresponding body parts.

The loss function of multipart contrastive loss can be represented as follows:
\begin{equation}
\mathcal{L}_{con}^{multi}=\frac{1}{N} \left(\mathcal{L}_{con}(\mathbf{S}_g,\mathbf{T}_g)+\mathcal{L}_{con}(\mathbf{S}_g,\mathbf{T}_s)+\sum_{i=1}^{P}\mathcal{L}_{con}(\mathbf{S}_{pi},\mathbf{T}_{pi})\right)
\end{equation}
where $N$ is the number of terms used to calculate the contrastive loss. The hierarchical approach offers distinct advantages by inherently respecting the compositional nature of sign language while enabling robust learning from partial observations. Through explicit modeling of part-whole relationships, it achieves superior generalization. As demonstrated in our experiments, this multi-granularity representation significantly outperforms global-only contrastive approaches while maintaining computational efficiency.

\subsection{Composite Training Objective}

Having introduced the individual components, we now formalize the complete optimization objective that jointly trains the skeleton encoder and text alignment modules.
The composite training objective combines classification and contrastive losses:

\begin{equation}
\mathcal{L}_{total} = \mathcal{L}_{cls}(\mathbf{S}_g) + \alpha\mathcal{L}_{con}^{multi}(\mathbf{S}, \mathbf{T})
\end{equation}
where $\mathcal{L}_{cls}$ denotes the standard cross-entropy classification loss, $\mathcal{L}_{con}^{multi}$ represents our multi-positive contrastive loss, and $\alpha$ is a fixed weighting parameter balancing the two objectives. The weighting parameter $\alpha$ controls the relative importance of semantic alignment versus classification accuracy, which we empirically set to 0.5 based on validation performance.

During training, the skeleton encoder processes input poses to generate $\mathbf{S}_g$ through the global average pooling of all joint nodes, while $\mathbf{S}_p$ is obtained by average pooling predefined groups of related joints. Both features are projected to match the text feature dimension via separate fully connected layers. The text descriptions generated by the GSP are encoded in fixed-dimensional representations $\mathbf{T}$ using the text encoder. At inference time, the method utilizes only the global features of the skeleton encoder $\mathbf{S}_g$ for the final prediction, ensuring no additional computational overhead compared to conventional skeleton-based approaches.
During inference, the text encoder and part-specific branches are discarded, reducing the computational graph to a single-stream skeleton encoder matching the efficiency of conventional approaches while retaining the benefits of multimodal training.
\section{Experiments}

\subsection{Experimental Setup}

\subsubsection{Datasets}

We conduct comprehensive evaluations on two large-scale sign language recognition datasets.

\textbf{SLR-500} \cite{huang2018attention} is a Chinese sign language dataset containing 500 vocabulary items performed by 50 native signers under controlled laboratory conditions. The dataset is divided into 87,500 training videos (35 signers) and 37,500 test videos (15 signers) for signer-independent evaluation. It provides comprehensive coverage of basic Chinese signs with natural variations in signing styles.

\textbf{AUTSL} \cite{sincan2020autsl} features 226 Turkish sign language gestures collected from 43 signers, with 28,142 training and 3,742 test samples. The dataset presents unique challenges due to significant inter-signer variations and diverse recording conditions. Its medium-scale vocabulary and realistic signing variations make it particularly valuable for robustness evaluation.
\subsubsection{Implementation details}

We employ HR-Net \cite{sun2019deep} to extract 87 semantically significant keypoints per frame, which our ablation studies identified as optimal for the representation of sign language. These keypoints are processed by a CTR-GCN \cite{chen2021channel} backbone with multiscale temporal convolution, preserving both spatial and temporal dynamics. The CLIP text encoder \cite{radford2021learning} processes multiple description variants, including global descriptions $T_g$, synonym variations $T_s$, and part-specific descriptions $T_p$, using a contrastive loss temperature parameter $\tau=0.1$ throughout all experiments. Training protocols vary by dataset: for SLR-500 we employ 110 training epochs with batch size 120, implementing a 5-epoch linear warm-up phase followed by an initial learning rate 0.06 (gradually annealed by ×0.1 at epochs 90 and 100) and weight decay 5e-4. The AUTSL dataset follows similar 110-epoch training, but with batch size 80, initial learning rate 0.04, and weight decay 4e-4 while maintaining identical reduction scheduling.

Our method consists of generated content in authoritative sign language resources, including three official Chinese dictionaries \cite{cadhh2018csl,Gu2018cnsl,Gu2019cma} and the Turkish National Sign Language Dictionary \cite{Kavak2015}. For text generation, we utilize Moonshot\footnote{\url{https://platform.moonshot.cn/}} AI models (v1-8K/v1-32K) to produce synonym variants, expert-verified descriptions, and part-specific texts. All implementations use PyTorch running on NVIDIA A100-PCIE-40GB GPU hardware.

\subsection{Comparison with State-of-the-Art Methods}
In this section, we compare our method with the existing state-of-the-art (SOTA) methods using the SLR-500 dataset. Additionally, we also conduct performance comparisons using the large-scale AUTSL sign language dataset.

\subsubsection{Performance Comparison on SLR-500 Dataset}

Current SOTA methods exhibit several limitations that our approach addresses. While SignBERT effectively models hand gestures as visual tokens and BEST successfully applies BERT-style pre-training to triplet units, these methods predominantly focus on manual features, potentially neglecting crucial non-manual elements like facial expressions and body postures. Similarly, MASA's motion-aware autoencoder, though powerful, faces an information bottleneck from single-positive contrastive learning.
\begin{table}[ht]
  \caption{Performance comparison on SLR-500 dataset.}
  \label{tab:performance-on-slr-500}
  \begin{tabular}{p{3.3cm}p{5cm}}
    \toprule
    \multicolumn{1}{l}{Method} &\multicolumn{1}{c}{Accuracy(\%)}\\
    \midrule
    ST-GCN \cite{yan2018spatial} & \multicolumn{1}{c}{90.0}\\
    SignBERT \cite{hu2021signbert} & \multicolumn{1}{c}{94.5}\\
    BEST \cite{zhao2023best} & \multicolumn{1}{c}{95.4}\\
    MASA \cite{zhao2024masa} & \multicolumn{1}{c}{96.3}\\
    SSRL \cite{zhao2024self} & \multicolumn{1}{c}{96.9}\\
    Ours (joint\textbackslash joint\_motion) & \multicolumn{1}{c}{96.01\textbackslash 95.28 }\\
    Ours (bone\textbackslash bone\_motion) & \multicolumn{1}{c}{94.37\textbackslash 94.19}\\
    Ours (4 streams fusion) & \multicolumn{1}{c}{\textbf{97.1}}\\
  \bottomrule
\end{tabular}
\end{table}

In contrast, our method demonstrates significant advantages in several aspects. First, our method benefits from the action descriptions of various body parts related to sign language in the text features, emphasizing various action features of the human body related to sign language. By incorporating these text features, our model can more comprehensively understand sign language actions, including not only hand gestures, but also other important information such as facial expressions and body postures. This helps the model capture subtle differences in sign language more accurately, achieving performance comparable to or even better than SOTA methods on the SLR-500 dataset. Furthermore, our method performs well in feature representation at both the joint and bone levels, indicating that the model can effectively utilize feature information at different levels. Particularly in the 4s setting, our method achieved an accuracy of 97.1\%, demonstrating its powerful capability to handle more complex sign language action sequences.

\subsubsection{Performance Comparison on AUTSL Dataset}
\begin{table}[t]
  \caption{Performance comparison on AUTSL dataset (Top-1 and Top-5 accuracy in \%).}
  \label{tab:performance-on-autsl}
  \centering
  \begin{tabular}{lcc}
    \toprule
    Method & Top-1 & Top-5 \\
    \midrule
    SL-TSSI-DenseNet \cite{laines2023isolated} & 93.13 & -- \\
    SSTCN \cite{jiang2021sign} & 93.37 & -- \\
    SAM-SLR-V1 \cite{jiang2021sign} & 95.45 & 99.25 \\
    AM-GCN-A \cite{liu2024asymmetric} & 96.27 & 99.48 \\
    SAM-SLR-V2 \cite{jiang2021skeleton} & 96.47 & 99.76 \\
    TMS-Net \cite{deng2024tms} & 96.62 & 99.71 \\
    SML \cite{deng2024sml} & 96.85 & 99.79 \\
    Ours & \textbf{97.07} & \textbf{99.89} \\
    \bottomrule
  \end{tabular}
\end{table}

To verify the generalization of our method, we also conducted experiments on the AUTSL dataset. The AUTSL dataset is known for its diversity and complexity, which poses greater challenges for SLR models. As shown in Table \ref{tab:performance-on-autsl}, our method also demonstrated excellent performance on the AUTSL dataset, achieving the highest Top-1 accuracy of 97.07\% and Top-5 accuracy of 99.89\%, representing a significant improvement over previous methods.

Performance improvement can be attributed to the multi-positive contrastive learning mechanism, which provides a boost in accuracy of 1.36\% by capturing inter-sign variations. Besides, our multimodal fusion strategy outperforms existing methods through comprehensive integration of manual and non-manual features. These results confirm our method's strong generalization across different sign languages and datasets, maintaining consistent performance advantages while handling AUTSL's inherent variability challenges.

\begin{table}[t]
  \caption{Effectiveness of GLM, KB, and optimized prompts in sign language recognition. VE: Visual Encoder.}
  \label{tab:component-effectiveness-study}
  \centering
  \begin{tabular}{cccc>{\centering\arraybackslash}p{2.2cm}}
    \toprule
    VE & LLM & KB & Optimized Prompt & Accuracy (\%) \\
    \midrule
    \checkmark & -- & -- & -- & 93.85 \\
    \checkmark & \checkmark & -- & -- & 93.57 \\
    \checkmark & \checkmark & \checkmark & -- & 94.89 \\
    \checkmark & \checkmark & \checkmark & \checkmark & \textbf{95.25} \\
    \bottomrule
  \end{tabular}
\end{table}
\subsection{Ablation Study}

\subsubsection{Combining Sign Language with LLMs}
\label{sec:component-effectiveness-study}
Our initial experiments revealed limitations when directly applying general purpose LLMs for the generation of sign language descriptions. The generated output contained substantial hallucinations, specifically fabricated action descriptions misaligned with actual signs, leading to a 0.27\% accuracy degradation in skeleton-text fusion tasks, as quantified in Table~\ref{tab:component-effectiveness-study}. To address these challenges, RAG is strategically utilized with domain-specific expert-validated sign language corpora to ensure description accuracy. Moreover, specialized prompts are crafted to filter non-action references (e.g., ``represents agreement'').

The ablation study demonstrates the contribution of each element: the base visual encoder achieves 93.85\% accuracy, while the raw LLM outputs degrade performance by -0.28\%. The integration of a knowledge base (KB) not only fully recovers but also substantially exceeds the baseline by +1.04\%. Furthermore, optimized prompts provide an additional improvement of +0.36\%. This systematic evaluation validates the effectiveness of our approach in addressing LLM hallucinations and enhancing the quality of action descriptions for sign language recognition.

\subsubsection{Multipart Contrastive Learning for Sign Language Recognition }

Our method enhances traditional classification-based SLR through a novel multipart contrastive learning mechanism. This approach effectively leverages LLM-generated knowledge (detailed action descriptions, synonyms, and expert grounded part-specific texts) to deepen the model's understanding of sign language semantics by aligning visual and textual representations.

\begin{table}[t]
  \caption{Ablation study of hierarchical multipart contrastive learning (based on joint data).}
  \label{tab:multipart-study}
  \centering
  \begin{tabular}{cccc>{\centering\arraybackslash}p{2.5cm}}
    \toprule
    VE & Synonym & Prompt & Multipart & Accuracy (\%) \tabularnewline
    \midrule
    \checkmark & -- & -- & -- & 93.85 \tabularnewline
    \checkmark & \checkmark & -- & -- & 94.50 \tabularnewline
    \checkmark & -- & \checkmark & -- & 94.93 \tabularnewline
    \checkmark & -- & -- & \checkmark & 95.25 \tabularnewline
    \checkmark & \checkmark & \checkmark & -- & 95.38 \tabularnewline
    \checkmark & \checkmark & \checkmark & \checkmark & \textbf{95.81} \tabularnewline
    \bottomrule
  \end{tabular}
\end{table}

\begin{table}[t]
  \caption{Comparison of the effects of different keypoint combinations in sign language recognition.}
  \label{tab:different-joint-sets}
  \centering
  \begin{tabular}{lcccc}
    \toprule
    Method & Num. Keypoints & Acc (\%) & Parts & +Multipart (\%) \\
    \midrule
    all & 133 & 59.22 & -- & -- \\
    base & 27 & 93.46 & 3 & 93.62 (0.16\%$\uparrow$) \\
    MASA & 49 & 93.66 & 3 & 94.01 (0.35\%$\uparrow$) \\
    Cosign & 76 & 93.57 & 5 & 94.52 (0.95\%$\uparrow$) \\
    Ours & 87 & 93.85 & 5 & \textbf{95.25} (\textbf{1.40\%}$\uparrow$) \\
    \bottomrule
  \end{tabular}
\end{table}
The experimental results show a successive improvement in performance with each additional element introduced. The inclusion of synonym variants contributes an increase of +0.65\%, while the use of optimized prompts adds another +1.08\% to the accuracy. Moreover, incorporating precise part-specific descriptions delivers the most substantial improvement of +1.40\%. When all these elements are fully integrated into the complete method, it achieves an impressive accuracy of 95.81\%. This validates the effectiveness of our contrastive learning strategy, which successfully bridges the gap between visual and textual representation spaces, captures the nuanced differences in actions, and combines expert knowledge with the capabilities of language models.
\subsubsection{Keypoint Selection Analysis}
\label{sec:keypoint-selection}
The selection of keypoint combinations significantly impacts the recognition performance in skeleton-based SLR. Table~\ref{tab:different-joint-sets} compares various approaches, demonstrating that our method achieves superior accuracy (93.85\%) with 87 keypoints, outperforming existing combinations (27-76 keypoints) while maintaining computational efficiency.

There are limitations to previous keypoint selection strategies. SAM-SLR \cite{jiang2021sign} significantly improved recognition accuracy by reducing the number of keypoints from 133 to 27. However, these methods have certain limitations in keypoint selection. SAM-SLR and MASA \cite{zhao2024masa} neglect facial, lip, and other non-hand information related to sign language; although CoSign \cite{jiao2023cosign} collects more keypoints, it still does not cover all relevant body parts. 

Through systematic examination of expert sign language descriptions, we identify 87 keypoints covering both manual (hands) and non-manual (face, mouth, body) locations commonly used in real-world signing. When combined with our multipart contrastive learning, this selection achieves a 1.4\% accuracy improvement, which is significantly higher than the gains of other methods (0.16-0.95\%). The results validate that complete coverage and textual grounding yield optimal recognition performance.

\subsection{Visualized Analysis}

To provide qualitative and deeper insights into our method's behavior, we conduct a comprehensive visual analysis using the SLR-500 trained model. This examination reveals how the integration of textual descriptions influences both spatial attention patterns and categorical performance across diverse sign categories.

\subsubsection{Visualization of Human Keypoint Attention Weights}

\begin{figure}[t]
  \centering
  \includegraphics[width=\linewidth]{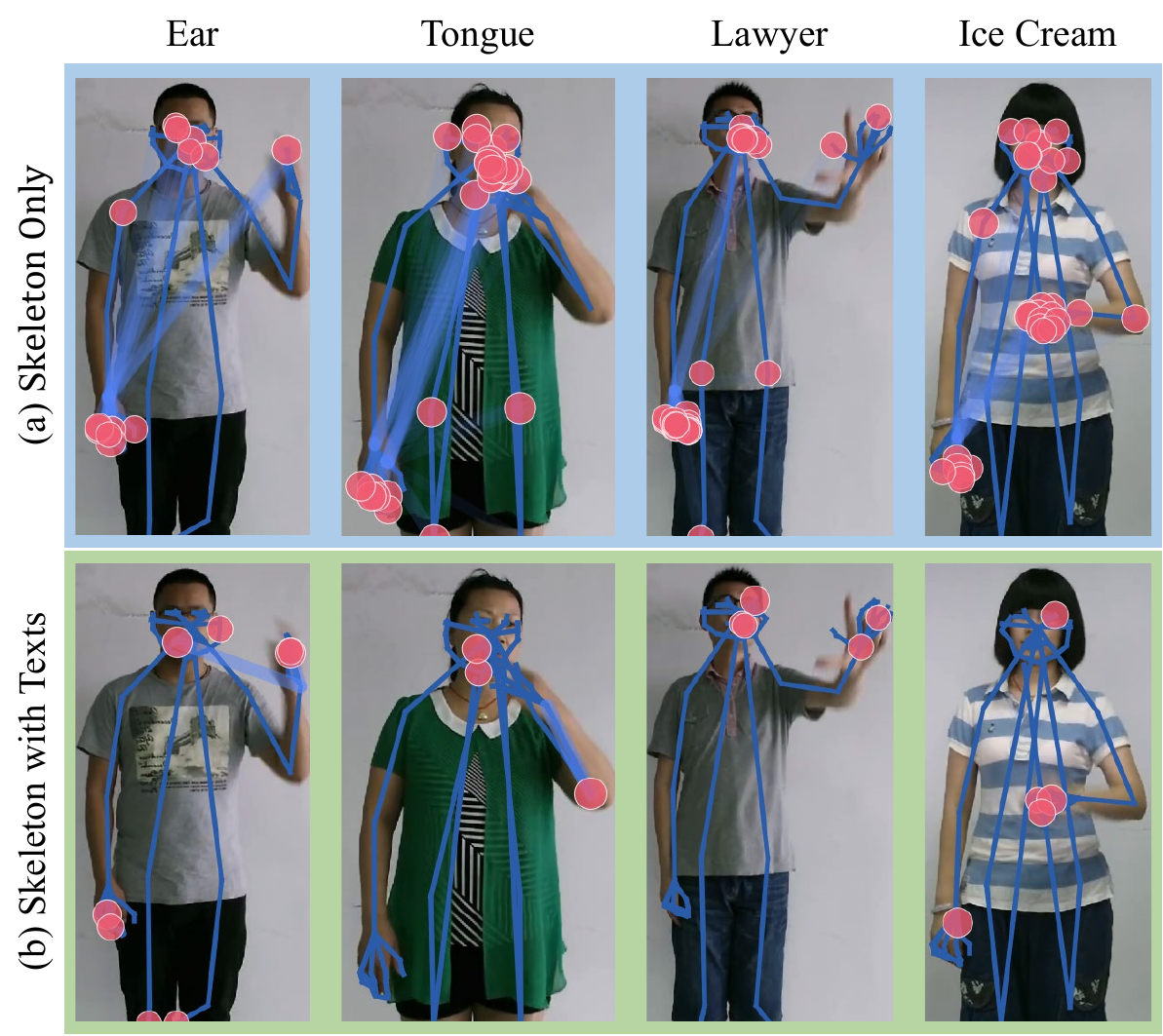}
\caption{Attention Visualization Comparison:  Skeleton-only (Top) baseline showing diffuse attention patterns. Our method's refined attention (Bottom) focusing on semantically relevant joints and relationships. Color coding: pink nodes=joint attention (size indicates importance), blue links=joint relationships (width/opacity indicate strength).}
  \label{fig:skeleton-visulazation}
  \Description{Comparative visualization of skeletal attention patterns showing the effect of text feature integration, with technical rendering details of joint importance and connection strength.}

\end{figure}

In Figure \ref{fig:skeleton-visulazation}, we visualize the attention patterns of our skeleton encoder to analyze the impact of contrastive learning based on multipart descriptions. The attention weights are derived from channel-averaged feature maps, where joint importance is computed through normalized attention aggregation. Before incorporating text features, the model exhibits diffuse attention across multiple joints, indicating weaker semantic alignment. After applying our MC method with text guidance, the model significantly reduces attention noise, focusing more precisely on linguistically meaningful regions. For instance, in Class ``Tongue'', the model correctly emphasizes left-hand and facial joints, while Class ``Ear'' shows enhanced attention to relevant inter-joint relationships. The visualization confirms that our method not only suppresses irrelevant keypoints but also strengthens the model’s ability to capture sign language semantics.

\subsubsection{Class-wise Performance Gains}

\begin{figure}[t]
  \centering
  \vspace{1cm}
  \includegraphics[width=\linewidth]{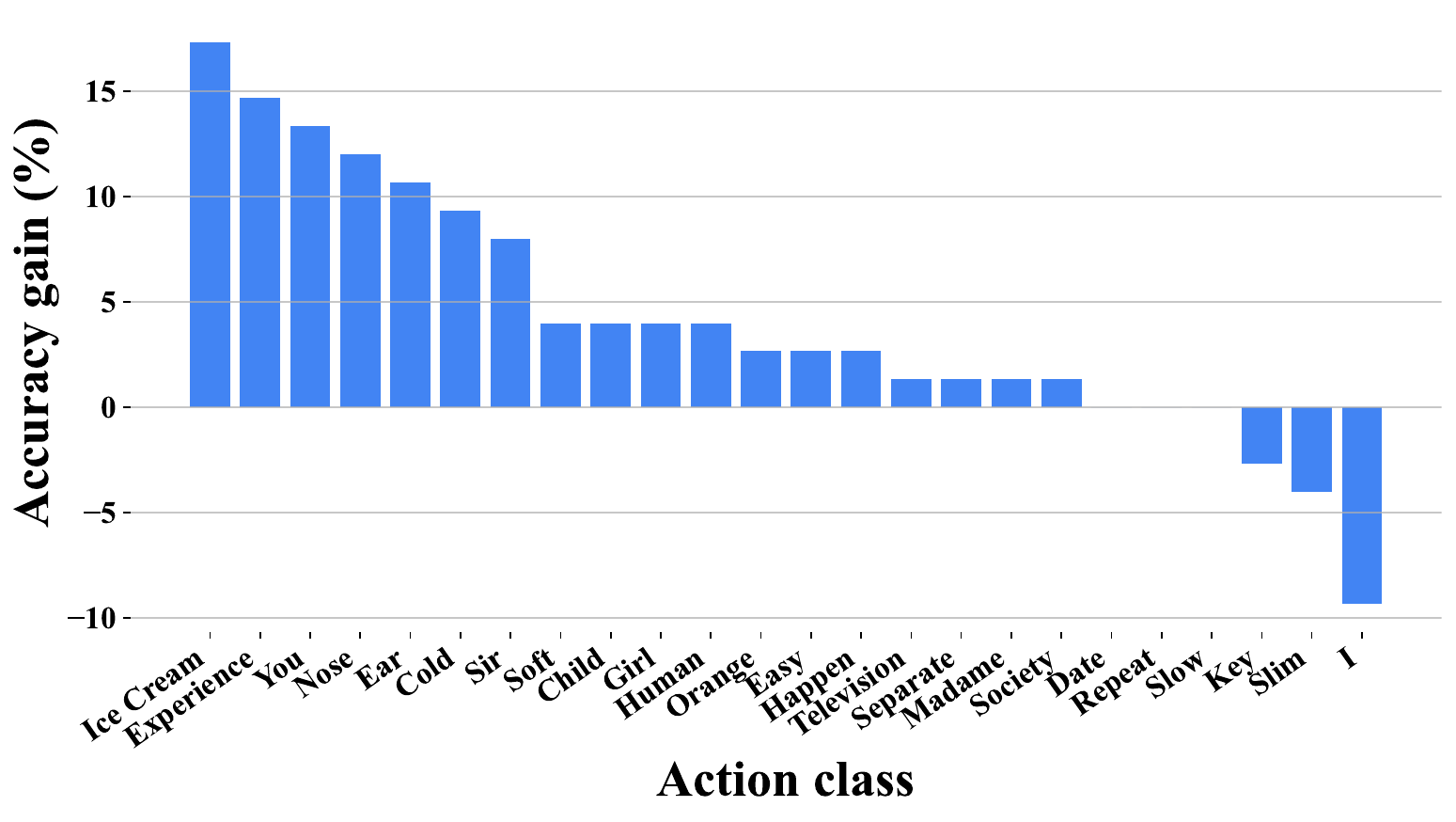}
  \caption{Class-wise Performance Gains}
  \label{fig:performance-gains}
  \Description{The class-wise gains obtained by the proposed MP-CGSDP method over the baseline ctr-gcn model on the SLR-500 dataset.}
\end{figure}

Figure~\ref{fig:performance-gains} presents a detailed analysis of the class-wise accuracy improvements of our GSP-MC method versus the baseline CTR-GCN model on SLR-500. The visual-text contrastive learning method produces markedly different impacts across sign categories, revealing important patterns about its operational mechanisms.

The substantial 17.33\% gain for ``ice cream'' stems from its precise description specifying ``bend fingers + press cheek + pinch thumb-index'', while ``experience'' improves by a notable 14.67\% with explicit ``circular motion near forehead'' trajectory details.
Conversely, ``thick'' declines 12\% due to its vague ``press down'' description lacking body part and trajectory specifics, and ``I'' drops 9.33\% when generated texts confuse pronouns. Recognition accuracy proves to be directly dependent on complete descriptions, particularly when specifying exact body configurations (``bend fingers''), locations (``cheek''), and motions (``circular'').

These results collectively demonstrate that our approach excels most for signs with concrete physical referents that permit clear part-specific decomposition. Performance variations directly correlate with description quality. The 4/5 of the classes maintaining or exceeding baseline performance confirms the overall robustness of the multipart contrastive learning approach.

\section{Conclusion}
This paper introduced GSP-MC, an innovative method for sign language recognition that combines the generative capabilities of LLM with advanced contrastive learning techniques. The Generative Sign-description Prompts leverage RAG to produce accurate, multipart sign descriptions grounded in expert knowledge. The Multi-positive Contrastive learning effectively aligns visual skeleton features with multiple textual representations of each sign, capturing nuanced variations in sign execution. Comprehensive evaluations on Chinese and Turkish sign language datasets demonstrated significant improvements over existing methods, achieving 97.1\% accuracy on SLR-500 and 97.07\% on AUTSL. The method's ability to model fine-grained sign dynamics while maintaining computational efficiency highlights its real-world potential for practical deployment. Future research directions include improving robustness to imperfect pose estimation and extending the approach to continuous sign language understanding.


\bibliographystyle{ACM-Reference-Format}
\bibliography{sample-base}








bv c


\end{document}